\documentclass[10pt,twocolumn,letterpaper]{article}
\usepackage[accsupp]{axessibility}
\usepackage[pagenumbers]{wacv}
\usepackage{graphicx}
\usepackage{amsmath}
\usepackage{amssymb}
\usepackage{booktabs}
\usepackage{color}
\usepackage{colortbl}
\usepackage{xcolor}
\usepackage{enumitem}
\usepackage{multicol}
\usepackage{multirow}
\usepackage{colortbl}
\usepackage{pifont}
\usepackage[linesnumbered,ruled,vlined]{algorithm2e}
\usepackage{babel} \usepackage[linesnumbered, ruled, vlined]{algorithm2e}
\SetKwInput{KwRequire}{Require}
\usepackage[pagebackref,breaklinks,colorlinks]{hyperref}
\definecolor{light_gray}{gray}{0.9}
\usepackage{blindtext}
\newcommand{\figref}[1]{Figure \ref{#1}}

\renewcommand{\paragraph}[1]{\vspace{1em}\noindent\textbf{#1}.}

\usepackage[capitalize]{cleveref}
\crefname{section}{Sec.}{Secs.}
\Crefname{section}{Section}{Sections}
\Crefname{table}{Table}{Tables}
\crefname{table}{Tab.}{Tabs.}

\DeclareMathOperator{\Tr}{Tr}

\begin{document}
\title{Layer-wise Auto-Weighting for Non-Stationary Test-Time Adaptation}
\author{Junyoung Park$^{1}$\quad Jin Kim$^{1}$\quad Hyeongjun Kwon$^{1}$\quad Ilhoon Yoon$^{1}$\quad Kwanghoon Sohn$^{1,2}$\thanks{Corresponding author} \\
$^{1}$Yonsei University\quad $^{2}$Korea Institute of Science and Technology (KIST) \\
{\tt\small \{jun\_yonsei, kimjin928, kwonjunn01, ilhoon231, khsohn\}@yonsei.ac.kr}
}
\maketitle
\begin{abstract}
Given the inevitability of domain shifts during inference in real-world applications, test-time adaptation (TTA) is essential for model adaptation after deployment.
However, the real-world scenario of continuously changing target distributions presents challenges including catastrophic forgetting and error accumulation.
Existing TTA methods for non-stationary domain shifts, while effective, incur excessive computational load, making them impractical for on-device settings.
In this paper, we introduce a layer-wise auto-weighting algorithm for continual and gradual TTA that autonomously identifies layers for preservation or concentrated adaptation.
By leveraging the Fisher Information Matrix (FIM), we first design the learning weight to selectively focus on layers associated with log-likelihood changes while preserving unrelated ones.
Then, we further propose an exponential min-max scaler to make certain layers nearly frozen while mitigating outliers.
This minimizes forgetting and error accumulation, leading to efficient adaptation to non-stationary target distribution.
Experiments on CIFAR-10C, CIFAR-100C, and ImageNet-C show our method outperforms conventional continual and gradual TTA approaches while significantly reducing computational load, highlighting the importance of FIM-based learning weight in adapting to continuously or gradually shifting target domains.
\footnote{Code is available at \url{https://github.com/junia3/LayerwiseTTA}}
\end{abstract}
\let\thefootnote\relax\footnotetext{This research was supported by the National Research Foundation of Korea (NRF) grant funded by the Korea government (MSIT) (NRF2021R1A2C2006703), and partly supported by the Institute of Information \& communications Technology Planning \& Evaluation (IITP) grant funded by the Korea government(MSIT) (No.2021-0-02068, Artificial Intelligence Innovation Hub).}
\begin{figure}[t]
    \centering
      \begin{subfigure}{1\linewidth}
          \includegraphics[width=1.0\linewidth]{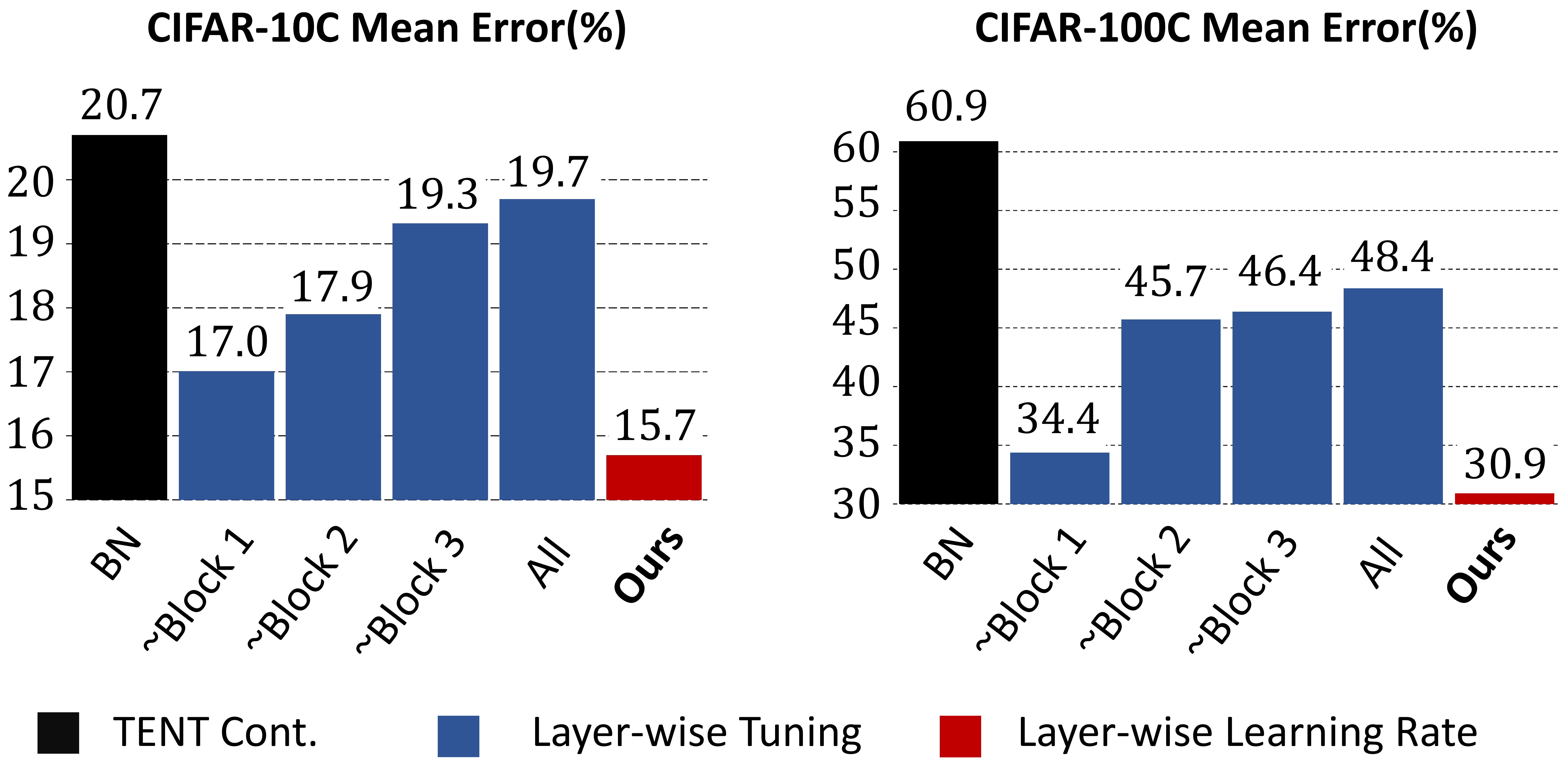}
          \caption{Mean error (\%) vs Tuning layer of TENT~\cite{TENT}}
          \label{fig:1a}
      \end{subfigure}
      \hfill
      \\
      \centering
      \begin{subfigure}{1\linewidth}
          \includegraphics[width=1.0\linewidth]{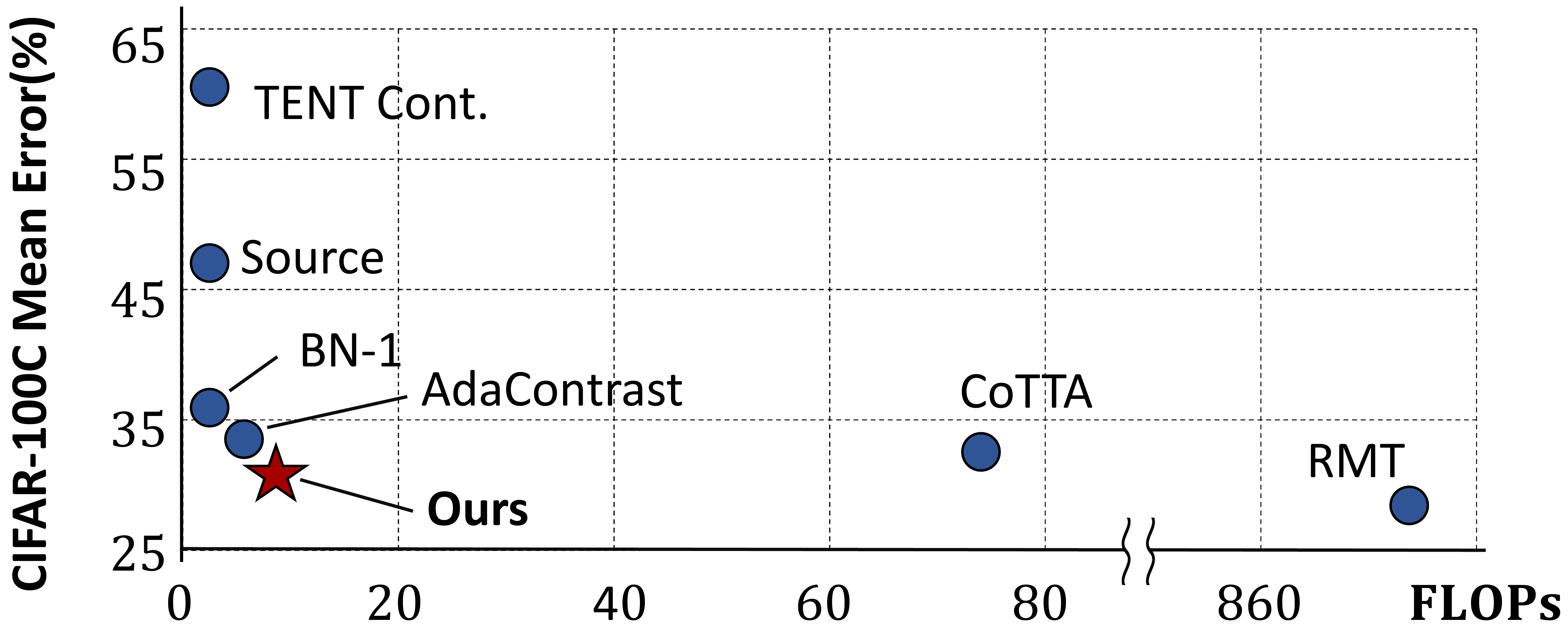}
          \vspace{-15pt}
          \caption{Mean error (\%) vs. FLOPs}
          \label{fig:1b}
      \end{subfigure}
      \hfill
      \\
      \vspace{-5pt}
      \caption{\textbf{(a) Continual TTA Performance between gradual layer tuning and Ours} in CIFAR-C dataset.
      While heuristically selecting layers to tune achieves favorable performance, our method outperforms through layer-wise auto-weighting.
      \textbf{(b) Adaptation performance of CTTA methods} on CIFAR-100C benchmarks.
      The $x-$ and $y-$axis are the FLoating point Operations Per second (FLOPs) and mean error (\%).
      Our method outperforms others with a significantly less computational load.}
      \label{fig:1}
      \vspace{-16pt}
\end{figure}

\section{Introduction}
\label{sec:intro}
Despite the recent advances in deep learning~\cite{resnet,dosovitskiy2020image,he2020momentum,he2022masked}, deep neural networks (DNNs) frequently encounter performance degradation when the source and target domains are different~\cite{long2016unsupervised,wei2021metaalign,kim2022pin}.
For instance, a pre-trained classification model suffers from this phenomenon when tested on corrupted images due to sensor deterioration.
Among various efforts to address these domain shifts, \textit{test-time adaptation} (TTA) has recently received significant attention, especially for its practicality~\cite{TENT,rmt}.
TTA aims to adapt the pre-trained source model by learning from unlabeled target data during inference where access to the source data is no longer available.
Since source data is unavailable during inference due to privacy concerns or legal constraints, TTA poses a realistic problem and is more challenging than unsupervised domain adaptation.
TTA can also be conducted in online scenarios where revisiting the past test samples is prohibited, adapting the model instantly using only the current test batch.
This makes TTA more applicable in on-device settings~\cite{ttt++, wei2021metaalign,gandelsman2022test}

Existing TTA methods often tackle the domain shift between the source and a fixed target domain by using pseudo labels or entropy regularization~\cite{BN, TENT}.
While these self-training methods have shown effectiveness, their improvements are demonstrated only in a single domain shift scenario at a time.
Since real-world target distribution tends to change continuously, it is necessary to consider the non-stationary target domain.
\cite{cotta} introduced \textit{continual test-time adaptation} (CTTA) where the model is adapted to a sequence of domain shifts.
The two challenges of CTTA are error accumulation by miscalibrated pseudo labels~\cite{guo2017calibration} and catastrophic forgetting by the gradual dilution of knowledge from the source pre-trained model.
To prevent the error accumulation, pioneering attempts~\cite{cotta, rmt} employ augmentation-averaged pseudo labels and an exponential moving averaged (EMA) model.
For source knowledge preservation, random parameter restore~\cite{cotta} and source prototypes~\cite{rmt} are used.
However, using these additional processes requires an excessive computational load than the original model as shown in \figref{fig:1b}.
This makes the methods less practical to use in on-device settings.

Despite the advancements in CTTA, previous works have not considered the heterogeneity of layers: each layer has a distinct role and captures different 
information~\cite{modulecritic1,modulecritic2,modulecritic3,gatys2016image}.
Recently, in the context of transfer learning, \cite{surgicalfinetune} identified certain layers of a pre-trained model that are already near-optimal for the target data through a heuristic approach.
They demonstrated that optimizing only the remaining layers helps preserve useful information from pre-training while enabling efficient learning for the target distribution.
In our preliminary CTTA experiment shown in ~\figref{fig:1a}, tuning the subset of layers improved adaptation capability compared to the vanilla or standard fine-tuning of the entropy minimization~\cite{TENT}.
We hypothesize that there is potential for further improvement of adaptation to continuously shifting target domains by identifying an optimal combination of tuning layers.

Inspired by this observation, we propose a novel layer-wise auto-weighting algorithm that autonomously identifies the layers that need preservation or concentrated adaptation.
The goal of this approach is to utilize the knowledge gained during source pre-training and efficiently adapt to the non-stationary target distribution.
To achieve this, we employ the Fisher Information Matrix (FIM) to approximate the second derivative of the log-likelihood~\cite{fisher}.
We calculate the FIM-based learning weight of each layer to measure the sharpness of the log-likelihood-layer parameters surface on target data~\cite{sharpness}.
Since higher sharpness indicates parameter sensitivity, we can identify layers to update or preserve.
Moreover, we introduce an exponential min-max scaler to amplify the learning weight difference across layers.
It makes certain layers nearly frozen while mitigating distortion of learning weights outliers.
As a result, our method can selectively focus on layers associated with log-likelihood changes in the target domain while preserving the unrelated ones.
Our method outperforms conventional CTTA approaches while significantly reducing computational load.
Experiments and ablation studies conducted on various benchmarks and networks, such as CIFAR-10C, CIFAR-100C, and ImageNet-C, prove the significance of the FIM-based learning weight.
Our key contributions are as follows:
\begin{itemize}
    \item We propose the layer-wise auto-weighted learning algorithm by leveraging the Fisher Information Matrix (FIM) to autonomously identify layers needing preservation or concentrated adaptation.
    \item We introduce an exponential min-max scaler to amplify the learning weight difference while mitigating distortion from outliers.
    \item We demonstrate comparable performance of our method while reducing computational load, compared to conventional approaches in both continual and gradual TTA on various benchmarks and networks.
\end{itemize}
\begin{figure*}[!ht]
    \centering
    \includegraphics[width=0.94\linewidth]{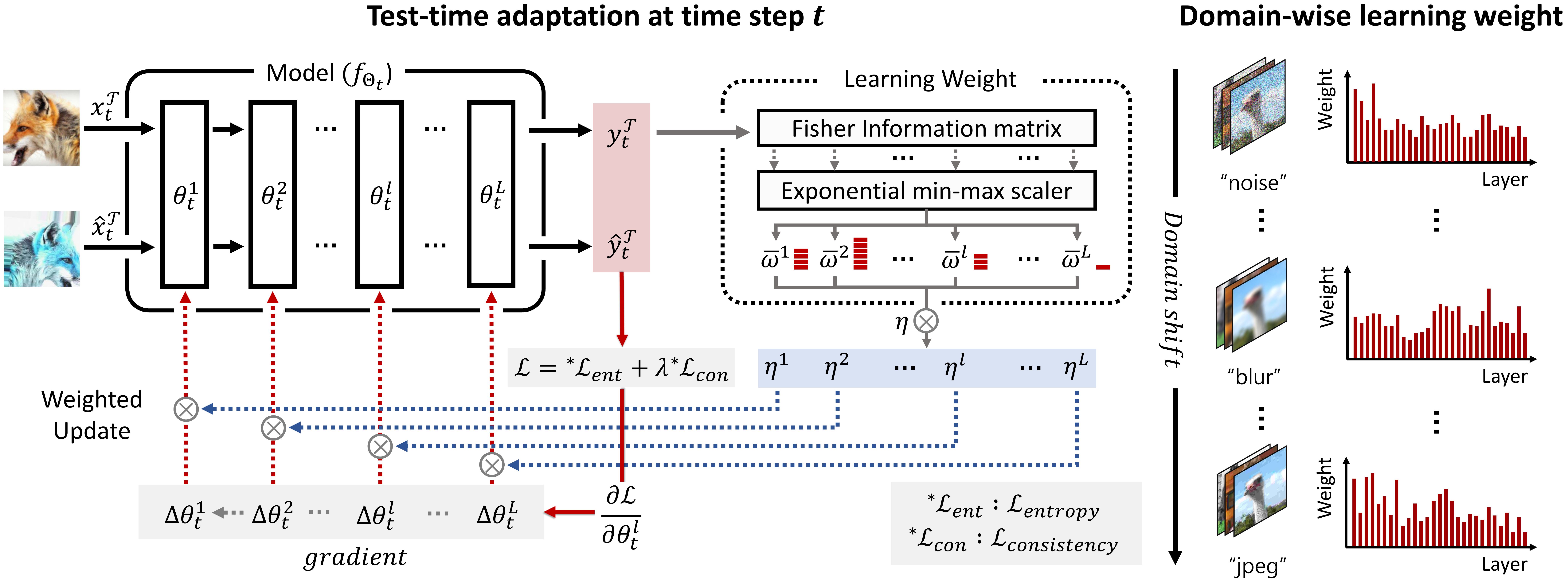}
    \vspace{-10pt}
    \caption{
\textbf{Layer-wise learning rate framework:} The model is adapted to current test batch $x_t^\mathcal{T}$.
Before updating the current model parameter, the Fisher Information Matrix (FIM) is empirically obtained from log-likelihood \wrt current batch images and update domain-level FIM.
We further employ an exponential min-max scaler to mitigate the distortion of learning weights from outliers. 
Finally, we can automatically update with adjusted learning weight and total objective $\mathcal{L}$ in Eq.~\eqref{eq:finalloss}}
\label{fig:framework}
\vspace{-12pt}
\end{figure*}

\section{Related Work}

\paragraph{Test-time adaptation} The goal of test-time adaptation (TTA) is to adapt source pre-trained models to the unlabeled target domain with no access to the source dataset.
Moreover, TTA can be considered an online manner where the predictions are needed immediately and the model is adapted using only the current test batch.
It has recently gained increasing attention~\cite{BN,TENT,ttt++,shot,improveTTA,LAME,zhang2022memo}.
The pioneering work, BN-1~\cite{BN} aligns the Batch Normalization (BN) statistics to the target domain while TENT~\cite{TENT} minimizes Shannon entropy~\cite{shannon1948mathematical} optimizing only the BN parameters.
Instead of updating only a portion of the network, Adacontrast~\cite{adacontrast} leverages the memory module with contrastive learning and updates the entire network.
Although previous TTA methods show significant improvement, they are only demonstrated for a single domain shift at a time.

\paragraph{Non-stationary Test-time Adaptation}
Typically, TTA operates under the assumption of a stationary scenario without specific constraints, but the real-world domain shifts are dynamic and constantly changing.
In this scenario, the non-stationary TTA settings can be categorized into two branches: Continual Test-Time Adaptation (CTTA) and Gradual Test-Time Adaptation (GTTA).
CoTTA~\cite{cotta}, the pioneering work of CTTA, was proposed to adapt continual domain shift in test-time.
They employ an exponential moving average (EMA) model to prevent error accumulation from miscalibration~\cite{guo2017calibration} and catastrophic forgetting.
In addition, RMT~\cite{rmt} leverages source prototypes for indirectly aligning the source distribution.
Moreover, gradual TTA~\cite{GTTA} assumes a gradual domain change in severity rather than an abrupt shift. 
GTTA~\cite{GTTA} employs style transfer to fill the gap between intermediate distribution shifts. 
However, they still suffer from heavy computational loads since using the auxiliary part like the EMA models.

\paragraph{Layer-wise fine-tuning} In transfer learning, preserving information from the source pre-training is important.
There are prominent approaches, such as partial parameter freezing~\cite{finetune1,finetune2,finetune3,finetune4,finetune5,finetune6}, and leveraging layer criticality refer to the loss surface~\cite{modulecritic1, modulecritic2, modulecritic3}.
Recently, surgical fine-tuning~\cite{surgicalfinetune} was proposed to heuristically find already near-optimal layers to the target domain and fine-tune only the remaining subset of all layers in the pre-trained models.
However, previous works addressed the case that the ground truth of target data is available.
In this paper, we propose a layer-wise auto-weighted adaptation method applicable to the unlabeled target data.

\paragraph{Fisher information matrix}
The Fisher information matrix (FIM) is a fundamental concept for defining intrinsic structures and performing gradient-based optimization.
From this perspective, Tan~\etal~\cite{fimrel} claims that FIM can achieve better results due to the adaptation of curvature information since it is a kind of second-order derivative.
Therefore, by conducting FIM in DNNs, FIM could adaptively adjust the step length throughout the training process.
In this regard, several works~\cite{EWC, onlineEWC} employ FIM to contiual learning to prevent catastrophic forgetting via adjusting the learning rate.
Despite the advantage of FIM, test-time adaptation with FIM, to the best of our knowledge, has not been demonstrated so far.
In this paper, we propose a simple but effective approach to adjust the learning rate by measuring the distribution shift via FIM.
\section{Proposed Method}
In real-world applications, the environment dynamically changes.
To effectively adapt to non-stationary test data, we propose an algorithm that autonomously assigns weights to layers needing preservation or concentrated adaptation.
Therefore, our method can adapt efficiently by preserving the source knowledge.
In this section, We revisit the problem statement of TTA in Sec.~\ref{section:3.1} and introduce the preliminaries and layer-wise weighted learning rate in Sec.~\ref{section:3.2}.
We also introduce the exponential min-max scaler that enables stable adaptation with main task loss in Sec.~\ref{section:3.3}.
Finally, we show our overall framework with consistency loss in Sec.~\ref{section:3.4}.
The overall procedure of our method is illustrated in \figref{fig:framework}.

\subsection{Problem statement}\label{section:3.1}
The goal of continual and gradual test time adaptation is to adapt the pre-trained model to a domain that is consistently changing in an online manner without access to the source dataset.
Given the model $f_{\Theta_0}$ where the pre-trained parameter $\Theta_0$ trained on the source domain $\mathcal{D}^\mathcal{S} = \{\mathcal{X}^\mathcal{S}, \mathcal{Y}^\mathcal{S}\}$, our goal is to adapt the model to the unlabeled target domain $\mathcal{D}^\mathcal{T} = \{\mathcal{X}^\mathcal{T}\}$, which consists of multiple domains that change sequentially.
At each time step $t$, the model $f_{\Theta_t}$ adapts to the target domain samples $x_t^\mathcal{T} \in \mathcal{X}^\mathcal{T}$, resulting in the updated model $f_{\Theta_{t+1}}$.
The model parameter $\Theta_t$ is composed of parameters $\theta^l_t$ for each layer $l$, such that $\Theta_t = \{\theta^{1}_t,\cdots,\theta^l_t,\cdots,\theta^{L}_t\}$.
Given the TTA loss function $\mathcal{L}$, the optimization of $\theta^l_t$ is as follows:
\begin{equation}\label{eq:1}
    \Delta \theta^l_t = \frac{\partial \mathcal{L}(x_t^\mathcal{T}; \Theta_t)}{\partial \theta^l_t},\quad
    \theta^l_{t+1} := \theta^l_t - \eta \ast \Delta \theta^l_t.
\end{equation}
where $\eta$ denotes the learning rate, which is typically set to be consistent across all layers.

\subsection{Layer-wise Learning Weight}\label{section:3.2}
Non-stationary domain adaptation aims to not only optimize effectively the target domain but also prevent error accumulation.
To achieve these objectives, we present the layer-wise learning weight to automatically adjust the learning rate $\eta$ depending on models and domain changes.
First, we have to infer the sharpness of the log-likelihood and the layer parameter $\theta^l_t$ surface by calculating the second derivative of the log-likelihood~\cite{sharpness}.
We can get the value of second derivatives through the Hessian matrix of the layer parameter $\theta^l_t$.
However, a direct computation of the Hessian matrix often fails on some large models~\cite{resnet, resnext, wideresnet} since the Hessian matrix increases quadratically $\mathbb{R}^{\lvert\theta^l_t \rvert}$ with the number of model parameters $\lvert\theta^l_t \rvert$~\cite{fimrel}.

\paragraph{Approximation of the Hessian matrix}
To mitigate computational overflow, we employ the Fisher information matrix (FIM).
The Hessian matrix requires two steps of derivatives.
However, the FIM can approximate the Hessian matrix of log-likelihood with one step of gradient~\cite{fisher}.
This distinction can help reduce the aforementioned problem, computational load.
To get FIM, we calculate log-likelihood from predicted output $p_{\Theta_t}(x_t^\mathcal{T})$ of the current test batch $x_t^\mathcal{T}$.
Then we can get the layer-wise score function $s(\theta^{l}_t; \cdot)$ which is the gradient of layer parameter $\theta^{l}_t$ from the log-likelihood as follows:
\begin{equation}\label{eq:score}
    s(\theta^l_t; x_t^\mathcal{T}) =\nabla_{\theta^l_t}{\log \left(p_{\Theta_t}(x_t^\mathcal{T})\right)}
\end{equation}
Then, the layer-wise FIM can be formulated as follows:
\begin{equation}\label{eq:layerwiseFIM}
    I_t^l=~\mathbb{E}_{x_t^\mathcal{T}}\left[ s(\theta^l_t; x_t^\mathcal{T}) s(\theta^l_t; x_t^\mathcal{T})^\top \right].
\end{equation}
Note that the layer-wise FIM $I_t^l$ is calculated for each layer $\theta^l_t$ with respect to the current test batch $x_t^\mathcal{T}$.
In our implementation, the layer-wise FIM is calculated based on the gradient of the layer computed using the negative log likelihood loss.
We further propose domain-level FIM $\tilde{I}_t^l$ from layer-wise FIM $I_t^l$ at time step $t$, inspired by~\cite{onlineEWC}. The domain-level FIM is formulated as:
\begin{equation}
    \tilde{I}_t^l = \tilde{I}_{t-1}^l + I_t^l
    \label{eq:domainlevelFIM}.
\end{equation}
Since the domain-level FIM $\tilde{I}_t^l$ is derived by accumulation rather than solely utilizing the current FIM, it includes more information about domain and model characteristics.
With the domain-level FIM, layer-wise learning weight can be obtained by taking a trace of FIM as follows:
\begin{equation}\label{eq:learningweight}
    w^l = \sqrt{\Tr(\tilde{I}_t^l)}.
\end{equation}

\paragraph{Layer-wise learning rate}
Based on the domain-level FIM in Eq.~\eqref{eq:domainlevelFIM}, we computed the learning weight $w^l$ for each layer.
The simplest way of layer-wise weighting is as follows:
\begin{equation}\label{eq:naiveweight}
\eta^l = \eta \ast w^l.
\end{equation}
Using this sharpness-based~\cite{sharpness} auto-weighting framework, the model efficiently adapts to the varying target domain by distributing layer-wise learning rates $\eta^l$.
\subsection{Exponential min-max scaler}\label{section:3.3}
However, the calculated learning weights are unbounded, which can deteriorate stable adaptation.
To realize this, we design an exponential min-max scaler to bound learning weights within the range of $0$ and $1$.
The exponential min-max scaler is employed by as follows:
\begin{equation}\label{eq:exponentiallr}
    \eta^l = \eta \ast \bar{w}^l,~~
    \bar{w}^l = \left(\frac{w^l - w^l_{\min}}{w^l_{\max} - w^l_{\min} + \epsilon}\right)^\tau,
\end{equation}
where $\tau$ is the exponential hyperparameter to mitigate the adverse effect of the outlier.
We set $\tau=1$ when the learning weights are scaled in a proper way. 
With this scaler, we can amplify the learning weight difference across layers making certain layers nearly frozen.
Moreover, it is more robust about the outlier value of the FIM than the vanilla min-max normalization.
Because when $\tau$ is greater than $1$, it reduces the over-scaled learning weights due to an excessively small minimum.
Conversely, as $\tau$ is lower than $1$, the under-scaled learning weights owing to a relatively large maximum can be boosted.
We can also mitigate the distortion of learning weights from outliers resulting in stable adaptation.
\subsection{Overall Update}\label{section:3.4}
\paragraph{Consistency Loss} Although layer-wise auto-weighted learning via FIM can reduce error accumulation on differential updates, there still exists the miscalibrated problem.
This miscalibration represents the difference between the information inherent in the data and the information the network acquires, due to inaccessibility to labels.
To address this problem, inspired from ~\cite{FixMatch,adacontrast}, we employ a self-training scheme in which the prediction of the original input batch $x_t^\mathcal{T}$ is considered as the pseudo label for the augmented batch $\hat{x}_t^\mathcal{T}$.
We further propose consistency loss $\mathcal{L}_\text{consistency}$ to regularize the network and prevent degradation from miscalibrated information.
The consistency loss $\mathcal{L}_\text{consistency}$ is formulated via conventional cross-entropy loss such that:
\begin{equation}\label{eq:consistencyloss}
    \mathcal{L}_\text{consistency} = -\mathbb{E}_{x_t^\mathcal{T}} \left[ \sum_{c=1}^C \sigma(y_t^\mathcal{T}) \log \sigma(\hat{y}_t^\mathcal{T}) \right],
\end{equation}
Where $y_t^\mathcal{T}$, $\hat{y}_t^\mathcal{T}$ are the corresponding logits and $\sigma(\cdot)$ is the sigmoid function.

\paragraph{Loss function}
Building upon negative log-likelihood, we compute the Fisher Information Matrix (FIM) and determine layer-wise weighted learning rates $\eta^l$.
Optimization operates batch-wise with derived learning rates.
Our overall loss function is composed of Shannon entropy loss $\mathcal{L}_\text{entropy}$ from the TENT~\cite{TENT} and consistency loss such that:
\begin{equation}\label{eq:finalloss}
    \mathcal{L} = \mathcal{L}_\text{entropy} + \lambda \mathcal{L}_\text{consistency},
\end{equation}
where $\mathcal{L}_\text{entropy}$ and $\mathcal{L}_\text{consistency}$ are the sum of each term for all samples in a batch, and the $\lambda$ controls the stability originated from the number of the class. 

\paragraph{Weighted Update}
Our full optimization framework is briefly visualized with \figref{fig:framework} and Algorithm~\ref{al_FIM}. 
The layer-wise auto-weighted learning procedure is summarized as follows:
(1) In the scenario of non-stationary test-time adaptation, the parameters of the previously trained network remain unchanged for the subsequent optimization phases.
The log-likelihood is computed using the predicted logits based on the prior of the previous model's parameters.
(2) Gradient computation is performed for each layer based on the log-likelihood and using the computed gradients, the empirical FIM is approximated as an estimation of the Hessian.
FIM is calculated online as a domain-level FIM according to Eq.~\eqref{eq:domainlevelFIM}.
(3) The layer importance, determined by characterizing the trace of the FIM, is used to adaptively adjust the learning rate for each parameter.
(4) According to the previously mentioned exponential min-max scaler Eq.~\eqref{eq:exponentiallr}, we updated each layer of our model with the gradient $\Delta \theta_t^l$ with respect to the loss function $\mathcal{L}$ in a layer-wise manner, as follows:

\begin{equation}\label{eq:layerwiseupdate}
    \theta^l_{t+1} := \theta^l_t - \eta^l \ast \Delta \theta^l_t.
\end{equation}
This allows for the separation of the importance of each learnable parameter embedded in each layer, enabling optimization with different step sizes.
\begin{algorithm}[t]
    \caption{Layer-wise auto-weighed learning}
    \label{al_FIM}
    \LinesNumbered
    \KwRequire{The encoder $f$, pretrained parameter $\Theta_0$, unlabeled target domain test dataset $\mathcal{X}^\mathcal{T}$}
    \For{each time step $t$}{
        Draw an \iid batch $x_t^\mathcal{T}\in\mathcal{X}^\mathcal{T}$,\\
        Compute score function $s(\theta^l_t; x_t^\mathcal{T})$ according to Eq.~\eqref{eq:score},\\
        Compute layer-wise Fisher Information Matrix $I_t^l$ per layer according to Eq.~\eqref{eq:layerwiseFIM},\\
        Update domain-level FIM according to Eq.~\eqref{eq:domainlevelFIM},\\
        Compute the learning weight of each layer using a trace of domain-level FIM according to Eq.~\eqref{eq:learningweight},\\
        Apply exponential min-max scaler to learning weight $w^l$ according to Eq.~\eqref{eq:exponentiallr},\\
        Update learning rate according to $\eta^l\gets{\eta \ast \bar{w}^l}$,\\
        Calculate gradient with objective function $\mathcal{L}$ in Eq.~\eqref{eq:finalloss},\\
        Update $\Theta_t$ to $\Theta_{t+1}$ by gradient descent with layer-wise learning rate in Eq.~\eqref{eq:layerwiseupdate}\\
        }
\end{algorithm}
\begin{table*}[ht]
    \caption{Online mean classification errors (\%) for the CIFAR-10C, CIFAR-100C and ImageNet-C continual test-time adaptation on the highest corruption severity. CIFAR-10C uses WideResNet-28, CIFAR-100C employs ResNeXt-29, and ImageNet-C utilizes ResNet-50. For a fair comparison, we recorded the performance of a single update or no update for all approaches.}
    \vspace{-7pt}
    \label{tab:continual}
    \centering\resizebox{\textwidth}{!}{
    \begin{tabular}{c|c|c|c|ccccccccccccccc|c|c}
    & Time & & & \multicolumn{15}{c|}{$t\xrightarrow{~~~~~~~~~~~~~~~~~~~~~~~~~~~~~~~~~~~~~~~~~~~~~~~~~~~~~~~~~~~~~~~~~~~~~~~~~~~~~~~~~~~~~~~~~~~~~~~~~~~~~~~~~~~~~~~~~~~~~~~~~~~~~~~~~~~~~~~~~~~~~~~~~~~~~~~~~~~~~~~~~~~~~~~~~~}$} & & \\ \toprule
    & Method & \rotatebox[origin=c]{90}{Source free} & \rotatebox[origin=c]{90}{Updates} & \rotatebox[origin=c]{70}{Gaussian} & \rotatebox[origin=c]{70}{Shot} & \rotatebox[origin=c]{70}{Impulse} & \rotatebox[origin=c]{70}{Defocus} & \rotatebox[origin=c]{70}{Glass} & \rotatebox[origin=c]{70}{Motion} & \rotatebox[origin=c]{70}{Zoom} & \rotatebox[origin=c]{70}{Snow} & \rotatebox[origin=c]{70}{Frost} & \rotatebox[origin=c]{70}{Fog} & \rotatebox[origin=c]{70}{Brightness} & \rotatebox[origin=c]{70}{Contrast} & \rotatebox[origin=c]{70}{Elastic} & \rotatebox[origin=c]{70}{Pixelate} & \rotatebox[origin=c]{70}{JPEG} & Mean & FLOPs \\ \midrule
    \multirow{7}{*}{\rotatebox[origin=c]{90}{CIFAR-10C}}
    & GTTA-MIX~\cite{GTTA} & \textcolor{red}{\ding{55}} & 1 & 26.0 & 21.5 & 29.7 & 11.1 & 30.0 & 12.2 & 10.5 & 15.1 & 14.1 & 12.3 &
    7.5 & 10.0 & 20.4 & 15.8 & 21.4 & 17.2 & 42.0 \\
    & RMT~\cite{rmt} & \textcolor{red}{\ding{55}} & 1 & \textbf{21.7} & \textbf{18.6} & \textbf{24.2} & \textbf{10.3} & \textbf{24.0} & \textbf{11.2} & \textbf{9.5} & \textbf{12.1} & \textbf{11.7} & \textbf{10.3} & \textbf{7.0} & \textbf{8.7} & \textbf{14.8} & \textbf{10.5} & \textbf{14.5} & \textbf{13.9} & 4252.7 \\ \cmidrule(lr){2-21}
    & Source~\cite{wideresnet} & \ding{51} & - & 72.3 & 65.7 & 72.9 & 46.9 & 54.3 & 34.8 & 42.0 & 25.1 & 41.3 & 26.0 & 9.3 & 46.7 & 26.6 & 58.5 & 30.3 & 43.5 & 10.5 \\
    & BN-1~\cite{BN} & \ding{51} & - & 28.1 & 26.1 & 36.3 & 12.8 & 35.3 & 14.2 & 12.1 & 17.3 & 17.4 & 15.3 & 8.4 & 12.6 & 23.8 & 19.7 & 27.3 & 20.4 & 10.5 \\
    & TENT-cont.~\cite{TENT} & \ding{51} & 1 & 24.8 & 20.6 & 28.6 & 14.4 & 31.1 & 16.5 & 14.1 & 19.1 & 18.6 & 18.6 & 12.2 & 20.3 & 25.7 & 20.8 & 24.9 & 20.7 & 10.5 \\
    & CoTTA~\cite{cotta} & \ding{51} & 1 & \textbf{24.3} & 21.3 & 26.6 & 11.6 & 27.6 & \textbf{12.2} & \textbf{10.3} & 14.8 & 14.1 & \textbf{12.4} & \textbf{7.5} & 10.6 & 18.3 & 13.4 & 17.3 & 16.2 & 357.0 \\
    & AdaContrast~\cite{adacontrast} & \ding{51} & 1 & 29.1 & 22.5 & 30.0 & 14.0 & 32.7 & 14.1 & 12.0 & 16.6 & 14.9 & 14.4 & 8.1 & \textbf{10.0} & 21.9 & 17.7 & 20.0 & 18.5 & 26.3 \\
    & \cellcolor{light_gray}\textbf{Ours} & \cellcolor{light_gray}\ding{51} & \cellcolor{light_gray}1 & \cellcolor{light_gray}24.5 & \cellcolor{light_gray}\textbf{18.9} & \cellcolor{light_gray}\textbf{25.1} & \cellcolor{light_gray}\textbf{11.6} & \cellcolor{light_gray}\textbf{26.9} & \cellcolor{light_gray}13.2 & \cellcolor{light_gray}10.4 & \cellcolor{light_gray}\textbf{14.2} & \cellcolor{light_gray}\textbf{13.4} & \cellcolor{light_gray}12.8 & \cellcolor{light_gray}8.4 & \cellcolor{light_gray}10.2 & \cellcolor{light_gray}\textbf{17.6} & \cellcolor{light_gray}\textbf{12.4} & \cellcolor{light_gray}\textbf{16.5} & \cellcolor{light_gray}\textbf{15.7$\pm$0.06} & \cellcolor{light_gray}42.0 \\ \midrule
    \multirow{7}{*}{\rotatebox[origin=c]{90}{CIFAR-100C}}
    & GTTA-MIX~\cite{GTTA} & \textcolor{red}{\ding{55}} & 1 & 39.4 & 34.4 & 36.6 & 24.7 & 36.8 & 26.6 & 24.3 & 30.1 & 28.9 & 34.6 & 22.8 & 25.1 & 30.7 & 26.9 & 34.7 & 30.4 & 8.62 \\
    & RMT~\cite{rmt} & \textcolor{red}{\ding{55}} & 1 & \textbf{37.4} & \textbf{33.8} & \textbf{34.3} & \textbf{24.8} & \textbf{32.0} & \textbf{25.3} & \textbf{23.6} & \textbf{26.2} & \textbf{26.2} & \textbf{28.9} & \textbf{21.9} & \textbf{23.5} & \textbf{25.4} & \textbf{23.2} & \textbf{27.4} & \textbf{27.6} & 873.72 \\ \cmidrule(lr){2-21}
    & Source~\cite{resnext} & \ding{51} & - & 73.0 & 68.0 & 39.4 & 29.3 & 54.1 & 30.8 & 28.8 & 39.5 & 45.8 & 50.3 & 29.5 & 55.1 & 37.2 & 74.7 & 41.2 & 46.4 & 2.16 \\
    & BN-1~\cite{BN} & \ding{51} & - & 42.1 & 40.7 & 42.7 & 27.6 & 41.9 & 29.7 & 27.9 & 34.9 & 35.0 & 41.5 & 26.5 & 30.3 & 35.7 & 32.9 & 41.2 & 35.4 & 2.16 \\
    & TENT-cont.~\cite{TENT} & \ding{51} & 1 & \textbf{37.2} & 35.8 & 41.7 & 37.9 & 51.2 & 48.3 & 48.5 & 58.4 & 63.7 & 71.1 & 70.4 & 82.3 & 88.0 & 88.5 & 90.4 & 60.9 & 2.16 \\
    & CoTTA~\cite{cotta} & \ding{51} & 1 & 40.1 & 37.7 & 39.7 & 26.9 & 38.0 & 27.9 & 26.4 & 32.8 & 31.8 & 40.3 & 24.7 & 26.9 & 32.5 & 28.3 & \textbf{33.5} & 32.5 & 73.34 \\
    & AdaContrast~\cite{adacontrast} & \ding{51} & 1 & 42.3 & 36.8 & 38.6 & 27.7 & 40.1 & 29.1 & 27.5 & 32.9 & 30.7 & 38.2 & 25.9 & 28.3 & 33.9 & 33.3 & 36.2 & 33.4 & 5.4 \\
    & \cellcolor{light_gray}\textbf{Ours} & \cellcolor{light_gray}\ding{51} & \cellcolor{light_gray}1 & \cellcolor{light_gray}39.1 & \cellcolor{light_gray}\textbf{34.2} & \cellcolor{light_gray}\textbf{36.1} & \cellcolor{light_gray}\textbf{25.3} & \cellcolor{light_gray}\textbf{36.2} & \cellcolor{light_gray}\textbf{27.0} & \cellcolor{light_gray}\textbf{25.1} & \cellcolor{light_gray}\textbf{30.7} & \cellcolor{light_gray}\textbf{29.2} & \cellcolor{light_gray}\textbf{35.9} & \cellcolor{light_gray}\textbf{24.4} & \cellcolor{light_gray}\textbf{26.9} & \cellcolor{light_gray}\textbf{31.3} & \cellcolor{light_gray}\textbf{27.5} & \cellcolor{light_gray}34.8 & \cellcolor{light_gray}\textbf{30.9$\pm$0.09} & \cellcolor{light_gray}8.62 \\ \midrule
    \multirow{7}{*}{\rotatebox[origin=c]{90}{ImageNet-C}} 
    & GTTA-MIX~\cite{GTTA} & \textcolor{red}{\ding{55}} & 1 & 80.5 & 74.7 & 72.4 & 77.8 & 75.7 & 64.3 & 54.0 & 57.0 & 58.6 & \textbf{44.6} & \textbf{33.9} & 67.5 & 49.4 & 44.7 & 49.3 & 60.3 & 32.98 \\
    & RMT~\cite{rmt} & \textcolor{red}{\ding{55}} & 1 & \textbf{77.3} & \textbf{73.2} & \textbf{71.1} & \textbf{73.1} & \textbf{71.2} & \textbf{61.2} & \textbf{53.7} & \textbf{54.3} & \textbf{58.0} & 46.1 & 38.2 & \textbf{58.5} & \textbf{45.4} & \textbf{42.3} & \textbf{44.5} & \textbf{57.9} & 1096.44 \\ \cmidrule(lr){2-21}
    & Source~\cite{resnet} & \ding{51} & - & 97.8 & 97.1 & 98.2 & 81.7 & 89.8 & 85.2 & 78.0 & 83.5 & 77.1 & 75.9 & 41.3 & 94.5 & 82.5 & 79.3 & 68.6 & 82.0 & 172.26 \\
    & BN-1~\cite{BN} & \ding{51} & - & 85.0 & 83.7 & 85.0 & 84.7 & 84.3 & 73.7 & 61.2 & 66.0 & 68.2 & 52.1 & \textbf{34.9} & 82.7 & 55.9 & 51.3 & 59.8 & 68.6 & 172.26 \\
    & TENT-cont.~\cite{TENT} & \ding{51} & 1 & 81.6 & 74.6 & 72.7 & 77.6 & 73.8 & 65.5 & 55.3 & 61.6 & 63.0 & 51.7 & 38.2 & 72.1 & 50.8 & 47.4 & 53.3 & 62.6 & 8.24 \\
    & CoTTA~\cite{cotta} & \ding{51} & 1 & 84.7 & 82.1 & 80.6 & 81.3 & 79.0 & 68.6 & 57.5 & 60.3 & 60.5 & 48.3 & 36.6 & \textbf{66.1} & \textbf{47.2} & \textbf{41.2} & \textbf{46.0} & 62.7 & 280.3 \\
    & AdaContrast~\cite{adacontrast} & \ding{51} & 1 & 82.9 & 80.9 & 78.4 & 81.4 & 78.7 & 72.9 & 64.0 & 63.5 & 64.5 & 53.5 & 38.4 & 66.7 & 54.6 & 49.4 & 53.0 & 65.5 & 20.6 \\
    & \cellcolor{light_gray}\textbf{Ours} & \cellcolor{light_gray}\ding{51} & \cellcolor{light_gray}1 & \cellcolor{light_gray}\textbf{80.4} & \cellcolor{light_gray}\textbf{73.8} & \cellcolor{light_gray}\textbf{71.2} & \cellcolor{light_gray}\textbf{77.5} & \cellcolor{light_gray}\textbf{72.9} & \cellcolor{light_gray}\textbf{63.9} & \cellcolor{light_gray}\textbf{54.1} & \cellcolor{light_gray}\textbf{57.9} & \cellcolor{light_gray}\textbf{59.8} & \cellcolor{light_gray}\textbf{46.3} & \cellcolor{light_gray}35.5 & \cellcolor{light_gray}67.2 & \cellcolor{light_gray}48.5 & \cellcolor{light_gray}44.9 & \cellcolor{light_gray}47.2 & \cellcolor{light_gray}\textbf{60.1$\pm$0.14} & \cellcolor{light_gray}32.98 \\ \bottomrule
    \end{tabular}
    }
    \vspace{-5pt}
\end{table*}
\section{Experiments}
We present the experimental results to introduce the effectiveness and efficiency of our proposed method.
We compare our method with state-of-the-art methods in different settings of non-stationary test-time adaptation: continual test-time adaptation (CTTA) and gradual test-time adaptation (GTTA), which results are presented Sec.~\ref{sec:cttares}, ~\ref{sec:gttares} respectively.
Then, we provide the results of extensive ablation studies and analysis in Sec.~\ref{sec:abl}.
\subsection{Experimental Settings}
\paragraph{Datasets and metrics} 
We evaluate the non-stationary test-time adaptation of our proposed method on CIFAR-10C, CIFAR-100C, and ImageNet-C, which are reconstructed from CIFAR-10~\cite{cifardata}, CIFAR-100~\cite{cifardata}, and ImageNet~\cite{imagenetdata} with prior shift corruption counterparts~\cite{corruptdata}. 
Each dataset contains an image set of $15$ corruption style including gaussian noise, shot noise, impulse noise, defocus blur, glass blur, motion blur, zoom blur, snow, frost, fog, brightness, contrast, elastic, pixelated, and jpeg.
We follow common protocol in~\cite{cotta, rmt}, evaluate on and report averaged classification error.

\begin{table*}[h]
\caption{Online mean classification errors (\%) for the CIFAR-10C, CIFAR-100C, and ImageNet-C gradual test-time adaptation. We present the performance results in two ways: by calculating the average performance across all severity levels (from level 1 to 5) and by calculating the average performance solely for the highest severity level (level 5). The number in brackets signifies the difference from the continual benchmark. For a fair comparison, we recorded the performance of a single update or no update for all approaches.}
\vspace{-8pt}
\label{tab:gradual}
\centering\resizebox{0.9\textwidth}{!}{
\begin{tabular}{c|c|cc|cccccc}
\toprule
& & GTTA-MIX & RMT & Source & BN-1 & TENT-cont. & AdaCont. & CoTTA & \cellcolor{light_gray}\textbf{Ours} \\
\midrule
& source-free & \textcolor{red}{\ding{55}} & \textcolor{red}{\ding{55}} & \ding{51} & \ding{51} & \ding{51} & \ding{51} & \ding{51} & \cellcolor{light_gray}\ding{51} \\ 
& Updates & 1 & 1 & - & - & 1 & 1 & 1 & \cellcolor{light_gray}1 \\ 
\midrule
\multirow{2}{*}{CIFAR-10C} & @level 1-5 & 10.5 & \textbf{8.1} & 24.7 & 13.7 & 20.4 & 12.1 & 10.9 & \cellcolor{light_gray}\textbf{9.6} \\
& @level 5 & 15.0 \footnotesize(-2.2) & \textbf{9.4} \footnotesize(-4.5) & 43.5 & 20.4 & 25.1 \footnotesize(+4.4) & 15.8 \footnotesize(-2.7) & 14.2 \footnotesize(-2.0) & \cellcolor{light_gray}\textbf{11.4} \footnotesize(-4.3) \\
\midrule
\multirow{2}{*}{CIFAR-100C} & @level 1-5 & 24.3 & \textbf{23.6} & 33.6 & 29.9 & 74.8 & 33.0 & 26.3 & \cellcolor{light_gray}\textbf{26.1} \\
& @level 5 & 27.6 \footnotesize(-2.8) & \textbf{24.3} \footnotesize(-3.2) & 46.4 & 35.4 & 75.9 \footnotesize(+15.0) & 35.9 \footnotesize(+2.5) & 28.3 \footnotesize(-4.2) & \cellcolor{light_gray}\textbf{28.2} \footnotesize(-2.7) \\
\midrule
\multirow{2}{*}{ImageNet-C} & @level 1-5 & 39.3 & \textbf{37.8} & 58.4 & 48.3 & 46.4 & 66.3 & 38.8 & \cellcolor{light_gray}\textbf{38.6} \\
& @level 5 & 51.8 \footnotesize(-8.5) & \textbf{40.2} \footnotesize(-17.7) & 82.0 & 68.6 & 58.9 \footnotesize(-3.7) & 72.6 \footnotesize(+7.1) & 43.1 \footnotesize(-19.6) & \cellcolor{light_gray}\textbf{41.6} \footnotesize(-18.5) \\
\bottomrule
\end{tabular}}
\vspace{-10pt}
\end{table*}

\paragraph{Implementation details}
For each dataset, we employ a different network, paired as follows: CIFAR-10C to WideResNet-28~\cite{wideresnet}, CIFAR-100C to ResNeXt-29~\cite{resnext}, and ImageNet-C to ResNet-50~\cite{resnet}. 
The networks are optimized in an online manner with batch sizes $200$, $200$, and $64$ and base learning rates $1e-3$, $5e-5$, $2e-4$, respectively. 
In our approach, we use the same Adam optimizer with $\beta = (0.9,~0.999)$ in all experiments. 
Hyper-parameter $(\lambda,~\tau)$ is set to $(0.1,~1.0)$ in CIFAR10-to-10C, $(1,~0.6)$ in CIFAR100-to-100C and $(10,~1.0)$ in ImageNet-to-ImageNet-C.

\subsection{Continual test-time adaptation}\label{sec:cttares} \paragraph{Experimental settings}
Similar to TTA setting used in TENT~\cite{TENT}, continual test-time adaptation (CTTA) departs from the source pre-trained model. However, while the TTA setting resets the model with source domain parameters after an adaptation to a single target domain, the CTTA considers multiple target domains and has the assumption that does not know when the domain changes. 
Therefore, the model is adapted to a sequence of test domains in an online manner. Following the CTTA benchmark in ~\cite{robustbench, rmt}, the sequence of corruptions consists of the highest severity level $5$ of all $15$ corruption domains.

\paragraph{Experimental results} 
Table~\ref{tab:continual} shows the results for corruption datasets in the continual setting with online measurement. 
The results show that our FIM-based method achieves comparable and better performance than the prior BN-based approach such as BN~\cite{BN}, and TENT-continual~\cite{TENT} with significant improvement.
Our proposed method shows more competitive performance and outperforms most prior works~\cite{adacontrast, cotta}.
The comparison to the most recent works~\cite{GTTA,rmt}, our method shows comparable performance despite our method being completely inaccessible to the source domain.
Specifically, RMT approach managed to alleviate this augmentation-induced bottleneck, but the computational overhead still exists due to the intricacies of the training framework itself, involving the use of source prototypes or the source replay process.
Moreover, GTTA-MIX introduced intermediate domain mixup methods and showed promising results on CTTA benchmarks, but they still need source datasets for acquiring additional information. 
To sum up, our proposed approach, requiring no additional frameworks and source information, enables optimization within a singular encoder. 
Notably, our proposed method achieves results of $15.7\%$, $30.9\%$, and $60.1\%$ for CIFAR-10C, CIFAR-100C, and ImageNet-C, respectively. 
These results are comparable to those obtained with computationally intensive methods such as CoTTA~\cite{cotta} and demonstrate performance on par with GTTA-MIX and RMT in non-source-free settings. For additional experimental results on model variation, please refer to Supplementary Sec.~\ref{sectionA}.
\begin{table}[htb!]
\caption{Comparison between AutoRGN and our method for mean classification error (\%). The number in brackets denotes the performance difference compared
to TENT continual~\cite{TENT}.}
\vspace{-5pt}
\label{tab:auto}
\centering\resizebox{0.95\linewidth}{!}{
\begin{tabular}{c|c|c|c}
\hline
\textbf{Method} & CIFAR-10C & CIFAR-100C & ImageNet-C \\
\hline\midrule
TENT cont.~\cite{TENT} & 20.7 & 60.9 & 62.6 \\ 
AutoRGN~\cite{surgicalfinetune} & 18.06 \footnotesize (-2.64) & 33.16 \footnotesize (-27.74) & 62.24 \footnotesize (-0.36) \\
\rowcolor{light_gray}\textbf{Ours} & \textbf{15.74} \footnotesize (-4.69) & \textbf{30.91} \footnotesize (-29.99) & \textbf{60.07} \footnotesize (-2.53) \\ \bottomrule
\end{tabular}
}
\vspace{-15pt}
\end{table}

\paragraph{Comparison with AutoRGN}
We further compare our approach with AutoRGN~\cite{surgicalfinetune} using the same backbone on CTTA benchmark datasets to ensure a fair comparison. The quantitative results are presented in Table~\ref{tab:auto}. AutoRGN outperforms TENT in the layer-wise learning approach of CTTA, achieving accuracy rates of $18.06\%$, $33.16\%$, and $62.24\%$ for CIFAR-10C, CIFAR-100C, and ImageNet-C. Although AutoRGN improves CTTA's performance compared to the baseline, our approach maintains a lower mean error.

\subsection{Gradual test-time adaptation}\label{sec:gttares}
\paragraph{Experimental settings} We also evaluate our proposed method on the gradual test-time adaptation (GTTA) benchmarks with mean classification error for CIFAR-10C, CIFAR-100C, and ImageNet-C datasets. 
While the CTTA online setting encounters the highest corruption domains sequentially, GTTA online setting encounters gradually increasing severity sequentially as follows: 

\noindent$\underbrace{\cdots \rightarrow 2 \rightarrow 1}_\text{Domain $t-1$} \xrightarrow[\text{shift}]{\text{domain}} \underbrace{1 \rightarrow 2 \rightarrow \cdots \rightarrow 5 \rightarrow \cdots \rightarrow 1}_{\text{Domain $t$, gradually changing severity}} \xrightarrow[\text{shift}]{\text{domain}}$.

\paragraph{Experimental results} 
We present the mean classification error results for GTTA in Table~\ref{tab:gradual}, encompassing all severity levels from $1$ to $5$, as well as reporting specifically on level $5$. 
Previous approaches, including TENT-continual~\cite{TENT} and AdaContrast~\cite{adacontrast}, exhibit degradation on specific datasets.
Although EMA baselines like CoTTA~\cite{cotta} and RMT~\cite{rmt} demonstrate performance enhancements for GTTA~\cite{GTTA}, the computational complexity and source dependency become bottlenecks during test-time adaptation. 
Our method, which employs layer-wise learning rates to update each parameter, yields competitive results with these approaches and notably outperforms the TENT continual baseline. 
Furthermore, our method shows improvement in mean error for severity level $5$, with $11.4\%$ on CIFAR10-to-10C, $28.2\%$ on CIFAR100-to-100C and $41.6\%$ on ImageNet-C datasets, compared to CoTTA and GTTA-MIX.

\subsection{Ablation study and Analysis}\label{sec:abl}
To further analyze and validate the components of our method, we conduct ablation experiments on three benchmark datasets in the continual test-time adaptation (CTTA) setting. Additionally, to demonstrate the impact of the Fisher information matrix across various domain shifts, we provide a graph depicting layer-wise learning weight.
\begin{figure*}[t]
    \centering
    \includegraphics[width=\linewidth]{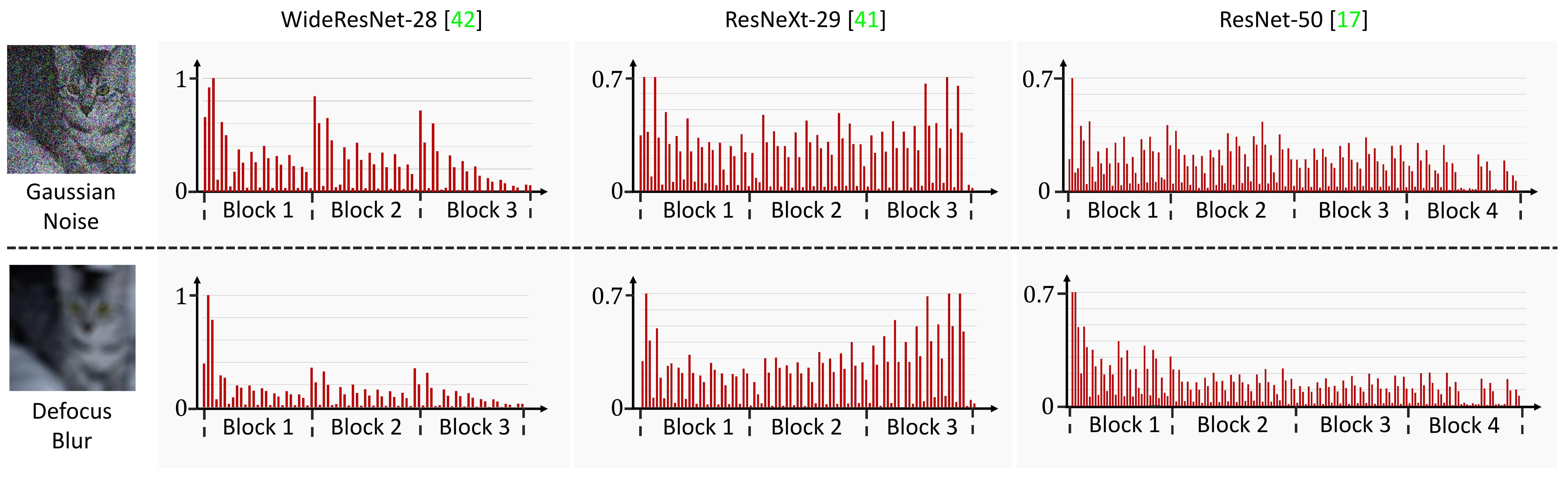}
    \vspace{-15pt}
    \caption{Results of ablation are presented, illustrating the normalized layer learning weight discrepancy between the dataset domain and each model. The first row displays the effects of Gaussian noise corruption at severity level 5, while the second row depicts Defocus blur corruption at severity level 5. The first column represents the CIFAR10-C dataset, the second column the CIFAR100-C dataset, and the last column the ImageNet-C dataset.
}
\label{fig:ablation}
\vspace{-15pt}
\end{figure*}

\paragraph{Contributions of objective}
We conduct additional experiments to validate the effect of the loss function in Eq.~\eqref{eq:finalloss} and its corresponding hyper-parameter $\lambda$, i.e., $\lambda \in \{0, 0.01, 0.1, 1.0, 10, 100\}$. 
In Table~\ref{tab:lambda}, we report the mean classification error rates on three corruption datasets with CTTA setting.
As we can see, the best performance is achieved at different values of $\lambda$, indicating that entropy minimization loss should be balanced with respect to the number of classes. 
In the case where $\lambda = 0$, entropy minimization is solely used as an optimization function and still shows improvement. 
For CIFAR-10C, solely optimized with entropy minimization and FIM learning weight improves performance about $3.5\%$ and with hyperparameter tuning, the best setting ($\lambda = 0.1$) improves performance about $4.9\%$ from the baseline~\cite{TENT}. In the case of CIFAR-100C, solely optimized with entropy minimization and FIM learning weight shows remarkable improvement in performance about $28.6\%$ and with hyperparameter tuning, the best setting ($\lambda = 1.0$) improves performance about $30.0\%$ from the baseline~\cite{TENT}. 
And for ImageNet-C, solely optimized with entropy minimization and FIM learning weight improves performance about $0.2\%$, and with hyperparameter tuning, the best setting ($\lambda = 10$) improves performance about $2.4\%$ from the baseline~\cite{TENT}. 
Here it indicates that our proposed FIM-based learning weight method shows improvements in all datasets even without the consistency loss in Eq.~\eqref{eq:consistencyloss} and with hyperparameter tuning for consistency loss function, we can further improve performance.
\begin{table}[h]
\caption{Mean classification error (\%) with varing $\lambda$.}
\vspace{-5pt}
\label{tab:lambda}
\centering\resizebox{1\linewidth}{!}{
\begin{tabular}{c|cccccc}
\hline
$\lambda$ & $0$ & $0.01$ & $0.1$ & $1.0$ & $10$ & $100$\\
\hline\midrule
CIFAR-10C & 17.22 & 16.65 & \cellcolor{light_gray}\textbf{15.74} & 17.76 & 19.04 & 17.93\\
CIFAR-100C & 32.29 & 32.40 & 32.23 & \cellcolor{light_gray}\textbf{30.91} & 31.66 & 31.74\\
ImageNet-C & 62.38 & 62.33 & 62.24 & 61.08 & \cellcolor{light_gray}\textbf{60.07} & 60.95\\ \bottomrule
\end{tabular}
}
\vspace{-15pt}
\end{table}

\paragraph{Exponential min-max scaler}
Also, we report the image classification error rate in Table~\ref{tab:importance} according to the scaler hyperparameter $\tau$, i.e., $\tau\in\{0.6, 0.7, 0.8, 0.9, 1.0, 1.1, 1.2\}$. 
From the results, we demonstrate that we can achieve remarkable progress in performance in each dataset by simply applying $\tau$ to learning weight in Eq.~\eqref{eq:exponentiallr}. 
For CIFAR-10C and ImageNet-C, $\tau = 1.0$ shows the best performance, which indicates that the original learning weight with min-max normalization is sufficient. 
Meanwhile for CIFAR-100C, $\tau = 0.6$ shows the best performance. We present additional $\tau$ variation results in Supplementary Sec.~\ref{sectionB2}.
\begin{table}[h]
\caption{Mean classification error (\%) with varing $\tau$.}
\vspace{-5pt}
\label{tab:importance}
\centering\resizebox{1\linewidth}{!}{
\begin{tabular}{c|ccccccc}
\hline
$\mathbf{\tau}$ & $0.6$ & $0.7$ & $0.8$ & $0.9$ & $1.0$ & $1.1$ & $1.2$\\
\hline\midrule
CIFAR-10C & 39.55 & 19.96 & 17.74 & 16.72 & \cellcolor{light_gray}\textbf{15.74} & 15.97 & 16.20 \\
CIFAR-100C & \cellcolor{light_gray}\textbf{30.91} & 31.42 & 32.16 & 32.42 & 32.76 & 33.03 & 33.30 \\
ImageNet-C & 77.29 & 69.89 & 63.44 & 61.08 & \cellcolor{light_gray}\textbf{60.07} & 61.21 & 61.65 \\ \bottomrule
\end{tabular}
}
\vspace{-15pt}
\end{table}

\paragraph{Layer-wise learning weight in domain shifts} 
In Figure~\ref{fig:ablation}, we show two corruption domains (gaussian noise, defocus blur) and their corresponding layer-wise learning weights using Eq.~\eqref{eq:exponentiallr} with $\tau = 1$. 
The learning weights differ across layers based on corruption types and networks.
Considering frequency components in corruption domains, variations appear in deeper layers of a hierarchical deep neural network (DNN). Domains with reduced high-frequency components, like defocus blur, tend to concentrate weights in initial layers. 
In contrast, noise-type domains, retaining higher frequency components, show more evenly distributed weights across network layers. 
These tendencies are apparent in domain-specific weight distributions.
These results demonstrate our method's robustness for test-time adaptation across diverse domains.
In Supplementary Sec.~\ref{sectionB3} and Sec.~\ref{sectionB4}, we provide visualizations of the diagonal FIM and additional discussion regarding layer-wise learning weights.
\section{Conclusion and Future Work}
This paper proposes a simple yet effective layer-wise auto-weighting algorithm that enhances non-stationary Test-Time Adaptation (TTA) performance. With considerably less computational load, our method is more suitable for edge devices requiring online adaptation.
First, we leverage the Fisher Information Matrix (FIM) to autonomously identify the layers that need preservation or concentrated adaptation. Second, we introduce an exponential min-max scaler to amplify the differences in learning weights across layers. This approach results in certain layers becoming nearly frozen while mitigating the distortion of learning weight outliers.
As a result, our method selectively focuses on layers associated with log-likelihood changes in the target domain while preserving the unrelated ones, minimizing catastrophic forgetting and error accumulation. Our FIM-based weighted learning leads to more efficient adaptation to non-stationary target domains.
Through extensive experiments and ablation studies on various benchmarks and networks, we have demonstrated that our method outperforms conventional CTTA approaches while significantly reducing the computational load. In this context, we believe that our work contributes to making test-time adaptation for edge devices more feasible in practice and hope that it will inspire further research in this area.

We also intend to employ our method in diverse recognition tasks, including segmentation and object detection, where a higher level of output granularity is needed.

\paragraph{Acknowledgement}
This research was supported by the Yonsei Signature Research Cluster Program of 2022 (2022-22-0002) and the KIST Institutional Program (Project No.2E31051-21-203).
{\small
\bibliographystyle{ieee_fullname}
\bibliography{egbib}
}
\clearpage
\appendix
In this document, we first complement the quantitative comparisons with encoder type variations (Sec.~\ref{sectionA}).
Moreover, we provide ablation studies (Sec.~\ref{sectionB}) including visualization of diagonals of Fisher Information Matrix (FIM) of our proposed method.
Our code is available at \url{https://github.com/junia3/LayerwiseTTA}.

\section{Additional Results}
\label{sectionA}
In this section, we conducted a comparative analysis between our proposed method and previous Continual Test-Time Adaptation (CTTA) approaches, including TENT~\cite{TENT}, BN-1~\cite{BN}, AdaContrast~\cite{adacontrast}, and CoTTA~\cite{cotta}. 

\subsection{Comparison with model variation}\label{sectionA1}
To validate the adaptive applicability of our proposed method across various encoder types, we conducted additional experiments.
We used several distinct encoder configurations for evaluation.
In the CIFAR-C benchmark, we used ResNext-29, WideResNet-28, WideResNet-40 from~\cite{robustbench}, and the ResNet-50 model from~\cite{ttt++}.
For simplicity, we refer to these models as RNXT29, WRN28, WRN40, and RN50, respectively.
However, the ImageNet-C~\cite{imagenetdata} pre-trained model parameters for RNXT29 and WRN are not available in the previous benchmark~\cite{robustbench}.
Therefore, we conducted experiments using pre-trained parameters from other benchmarks for ResNext-50~\cite{resnext} and WideResNet-50~\cite{wideresnet} models.
To simplify, we refer to these models as RNXT50 and WRN50, respectively. Although the pre-trained ResNet-50~\cite{resnet} for ImageNet-C differs from the one in~\cite{ttt++}, we will use the term RN50 for convenience.

\paragraph{Comparison on CIFAR-10C} 
Table~\ref{tab:cifar10c} presents the average classification errors on CIFAR-10C~\cite{cifardata} with the CTTA setting using four distinct encoder configurations.
Our method demonstrates an average mean error of $9.8\%$, $15.7\%$, $11.0\%$, and $12.8\%$ in each model framework, surpassing all previous source-free TTA methods across all encoder architectures.
Previously proposed methods, such as TENT~\cite{TENT}, CoTTA~\cite{cotta} and AdaContrast~\cite{adacontrast}, exhibited favorable performance in specific network structures but struggled to achieve stable adaptation performance across all network architectures.
In contrast, our method consistently delivers strong performance, regardless of the network structure, confirming its effectiveness.
\begin{table}[htb!]
\caption{Classification mean error (\%) for the CIFAR-10C online CTTA task on the highest corruption severity level 5. We report the performance of each method averaged over 5 runs.}
\label{tab:cifar10c}
\centering\resizebox{\linewidth}{!}{
\begin{tabular}{c|c|c|c|c}
\toprule
Method & RNXT29~\cite{robustbench} & WRN28~\cite{robustbench} & WRN40~\cite{robustbench} & RN50~\cite{ttt++} \\ \midrule\midrule
Source & 18.0 & 43.5 & 18.3 & 48.8 \\
BN-1~\cite{BN} & 13.3 & 20.4 & 14.6 & 16.1 \\
TENT-cont.~\cite{TENT} & 14.8 & 20.7 & 12.5 & 14.8 \\
CoTTA~\cite{cotta} & 11.0 & 16.2 & 12.7 & 13.1 \\
AdaContrast~\cite{adacontrast} & 11.0 & 18.5 & 11.9 & 14.5 \\
\cellcolor{light_gray} \textbf{Ours} & \cellcolor{light_gray}\textbf{9.8} & \cellcolor{light_gray}\textbf{15.7} & \cellcolor{light_gray}\textbf{11.0} & \cellcolor{light_gray}\textbf{12.8}\\
\bottomrule
\end{tabular}
}\vspace{-10pt}
\end{table}

\paragraph{Comparison on CIFAR-100C}
Additionally, we conducted evaluations on CIFAR-100C~\cite{cifardata} using the CTTA setting with our proposed method.
In Table~\ref{tab:cifar100c}, we can observe similar issues as with previous approaches in terms of model variations. 
TENT~\cite{TENT} shows improved performance in WRN40 and RN50 except for RNXT29 than BN-1, which simply updates batch normalization statistics.
Regarding CoTTA~\cite{cotta} and AdaContrast~\cite{adacontrast}, they show improvement in RNX29 and RN50 but exhibit performance degradation in WRN40.
Our method demonstrates outstanding performance across all variations, confirming its model-agnostic usability. 
It achieves $30.9\%$, $35.0\%$, and $36.2\%$ with each encoder.
\begin{table}[htb!]
\caption{Classification mean error (\%) for the CIFAR-100C online CTTA task on the highest corruption severity level 5. We report the performance of each method averaged over 5 runs.}
\label{tab:cifar100c}
\centering\resizebox{\linewidth}{!}{
    \begin{tabular}{c|c|c|c}
    \toprule
    Method & RNXT29~\cite{robustbench} & WRN40~\cite{robustbench} & RN50~\cite{ttt++} \\ \midrule\midrule
    Source & 46.4 & 46.8 & 73.8 \\
    BN-1~\cite{BN} & 35.4 & 39.3 & 43.7 \\
    TENT-cont.~\cite{TENT} & 60.9 & 36.9 & 44.2 \\
    CoTTA~\cite{cotta} & 32.5 & 38.2 & 37.6 \\
    AdaContrast~\cite{adacontrast} & 33.4 & 37.1 & 41.3 \\
    \cellcolor{light_gray} \textbf{Ours} & \cellcolor{light_gray}\textbf{30.9} & \cellcolor{light_gray}\textbf{35.0} & \cellcolor{light_gray}\textbf{36.2}\\
    \bottomrule
    \end{tabular}
    }
\end{table}

\paragraph{Comparison on ImageNet-C}
We also conducted evaluations on ImageNet-C~\cite{imagenetdata} with CTTA setting using our proposed method. 
Although BN-1~\cite{BN}, which does not require optimization, improved performance compared to the source approach, its performance significantly declined compared to optimization-based approaches.
In this experiment, TENT~\cite{TENT} outperforms CoTTA~\cite{cotta} and AdaContrast~\cite{adacontrast}, both of which update the entire set of parameters. 
TENT reports $62.6\%$, $58.7\%$, and $57.7\%$ accuracy in each architecture. 
Nevertheless, we demonstrate that our method is superior to others in all model frameworks.
Our proposed method achieves accuracy rates of $60.1\%$, $57.5\%$, and $56.4\%$, improving by $2.6\%$, $2.3\%$, and $1.3\%$ over TENT, respectively.
\begin{table}[htb!]
\caption{Classification mean error (\%) for the ImageNet-C online CTTA task on the highest corruption severity level 5. We report the performance of each method averaged over 5 runs.}
\label{tab:imagenetc}
\centering\resizebox{\linewidth}{!}{
    \begin{tabular}{c|c|c|c}
    \toprule
    Method & RN50~\cite{resnet} & RNXT50~\cite{resnext} & WRN50~\cite{wideresnet} \\ \midrule\midrule
    Source & 82.0 & 78.9 & 78.9 \\
    BN-1~\cite{BN} & 68.6 & 67.1 & 66.2 \\
    TENT-cont.~\cite{TENT} & 62.6 & 58.7 & 57.7 \\
    CoTTA~\cite{cotta} & 62.7 & 59.8 & 57.9 \\
    AdaContrast~\cite{adacontrast} & 65.5 & 63.1 & 63.3 \\
    \cellcolor{light_gray} \textbf{Ours} & \cellcolor{light_gray}\textbf{60.1} & \cellcolor{light_gray}\textbf{57.5} & \cellcolor{light_gray}\textbf{56.4}\\
    \bottomrule
    \end{tabular}
    }
\end{table}

\section{Ablation study}
\label{sectionB}
\subsection{Ablations with domain-level FIM}\label{sectionB1}
In all experimental settings, we used domain-level Fisher Information Matrix (FIM), which accumulates information from continuous domain samples, rather than solely relying on temporal FIM.
To verify the effectiveness of our method, we further introduced a hyperparameter $\gamma$ for the domain-level FIM in Eq.~\eqref{eq:domainlevelFIM} of our main paper, inspired by~\cite{onlineEWC}:
\begin{equation}
    \tilde{I}_t^l = \gamma\tilde{I}_{t-1}^l + I_t^l.
    \label{eq:domainlevelFIM_ablation}
\end{equation}
In Table~\ref{tab:domain_ablation}, we compared the performance of CTTA by varying the $\gamma$ of the regulated version of FIM from 0 to 1.
Since $\gamma$ is set to be less than 1, we can adjust our domain-level FIM from the batch level to the domain level.
For CIFAR-10C~\cite{cifardata}, our method achieved an accuracy of $15.74\%$, representing a $0.62\%$ improvement over the performance when $\lambda = 0$.
Our method exhibits an improvement of $0.86\%$ for CIFAR-100C~\cite{cifardata} and $0.87\%$ for ImageNet-C~\cite{imagenetdata}. This demonstrates that our domain-level FIM outperforms the FIM which focus on short intervals.
\begin{table}[htb!]
\caption{Mean classification error (\%) with varing $\gamma$.}
\label{tab:domain_ablation}
\centering\resizebox{\linewidth}{!}{
    \begin{tabular}{c|cccc|c}
    \toprule
    $\gamma$ & 0 & 0.3 & 0.6 & 0.9 & \cellcolor{light_gray}\textbf{Ours} \\ \midrule\midrule
    CIFAR-10C & 16.36 & 16.24 & 16.01 & 15.95 & \cellcolor{light_gray}\textbf{15.74}\\
    CIFAR-100C & 31.77 & 31.74 & 31.70 & 31.58 & \cellcolor{light_gray}\textbf{30.91} \\
    ImageNet-C & 60.94 & 60.84 & 60.79 & 60.67 & \cellcolor{light_gray}\textbf{60.07} \\ \bottomrule
    \end{tabular}
}
\end{table}

\begin{figure}[b!]
    \centering
    \includegraphics[width=\linewidth]{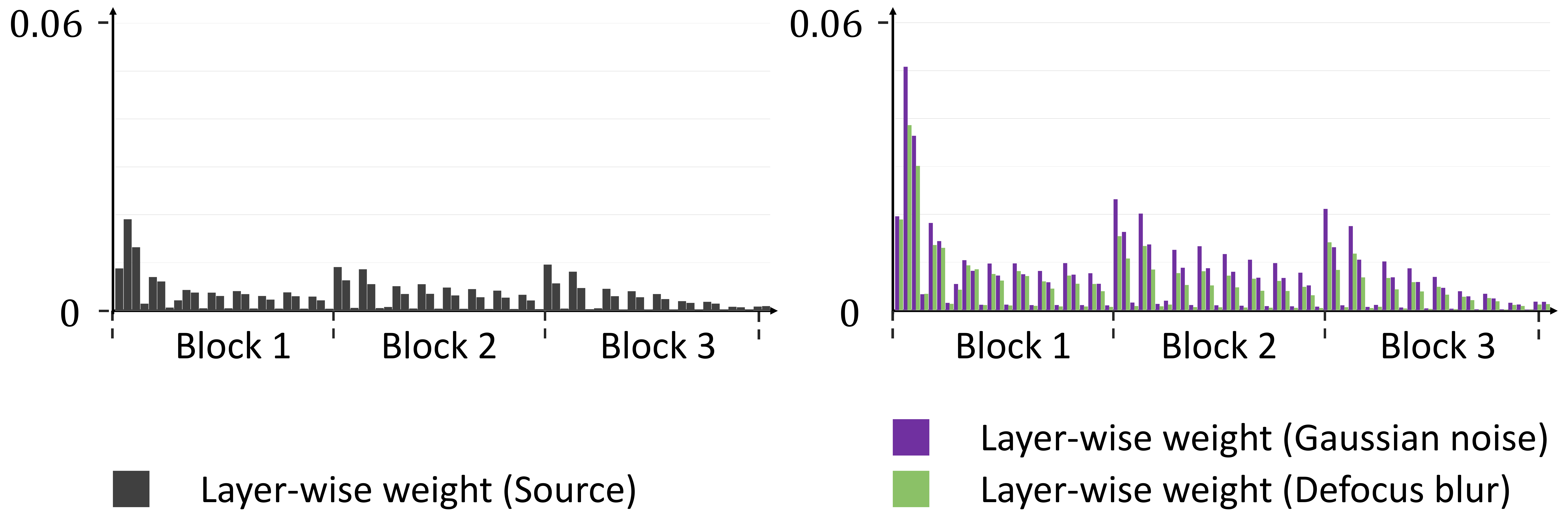}
    \caption{Layer-wise weights comparison between the source and target domains.}
    \label{fig:ablation_supplementary}
\end{figure}

\begin{figure}[t!]
    \centering
    \includegraphics[width=\linewidth]{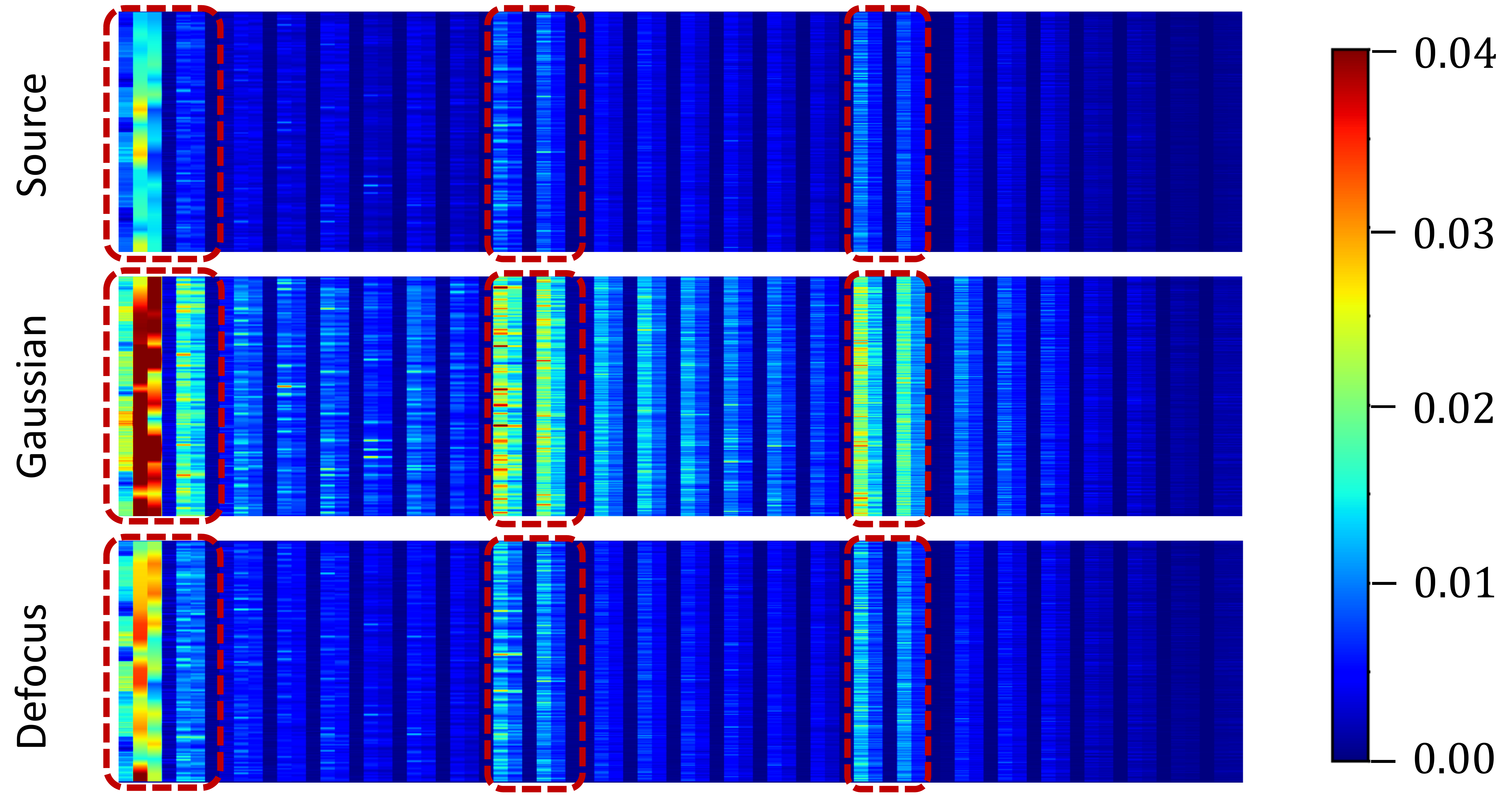}
    \caption{\textbf{Diagonal of the FIM per layer} with the source domain, gaussian noise domain and defocus blur domain in the CIFAR-10C dataset. The x- and y-axes represent the layer index and the diagonal values of the layer-wise FIM, respectively. This demonstrates that our method is robust to any kind of domain distribution.}
    \label{fig:fim_diag}
\end{figure}
\begin{table*}[t]
\caption{Mean classification error (\%) with varying $\tau$.}
\label{tab:moretau}
\centering\resizebox{0.9\linewidth}{!}{
\begin{tabular}{c|ccccccccccccc}
\toprule
$\mathbf{\tau}$ & $0.0$ & $0.1$ & $0.2$ & $0.3$ & $0.4$ & $0.5$ & $0.6$ & $0.7$ & $0.8$ & $0.9$ & $1.0$ & $1.1$ & $1.2$\\
\midrule\midrule
CIFAR-10C & 89.30 & 89.35 & 88.85 & 86.03 & 78.28 & 65.92 & 39.55 & 19.96 & 17.74 & 16.72 & \cellcolor{light_gray}\textbf{15.74} & 15.97 & 16.20 \\
CIFAR-100C & 37.16 & 33.98 & 32.77 & 32.12 & 31.83 & 31.75 & \cellcolor{light_gray}\textbf{30.91} & 31.42 & 32.16 & 32.42 & 32.76 & 33.03 & 33.30 \\
ImageNet-C & 99.57 & 99.39 & 98.72 & 97.10 & 94.79 & 90.95 & 77.29 & 69.89 & 63.44 & 61.08 & \cellcolor{light_gray}\textbf{60.07} & 61.21 & 61.65 \\ \bottomrule
\end{tabular}
}
\end{table*}

\subsection{Ablations with exponential min-max scaler}\label{sectionB2}
In addition to our main paper, we show additional ablations on the hyperparameter $\tau$ of the exponential min-max scaler in Eq.~\eqref{eq:exponentiallr} of our main paper. Table~\ref{tab:moretau} provides additional results for $\tau$ values ranging from $0$ to $1.2$ for each dataset in the CTTA benchmark. When $\tau = 1$, it represents the result obtained by applying simple min-max normalization to the weight importance from domain-level FIM. Conversely, for $\tau = 0$, it assumes that all weight importance values have equal weights of $1$. As $\tau$ approaches zero, layer-wise weights tend to have relatively uniform values, and the overall performances diverge during the adaptation process.

\subsection{Layer-wise learning weights in each domain}\label{sectionB3}
In the main paper, we attempted to apply layer-wise learning rates according to the domain shift problem. 
From this perspective, it should also be noted that if there is no domain shift, the learning weights will assume a minimal value.
To demonstrate this, we compare the learning weights from the source domain with those from the target domain.
Specifically, for the target domain, we choose gaussian noise and defocus blur for comparison.
In Figure~\ref{fig:ablation_supplementary}, we compare the learning weights measured by the pre-trained model on the CIFAR10-to-10C dataset without applying the exponential min-max normalization in Eq.~\eqref{eq:exponentiallr} of our main paper. 
The learning weights for gaussian noise/defocus blur vary by layer depending on the type of corruption, demonstrating that our model can adapt its learning rate to different types of domain shifts.
On the other hand, the learning weights in the source domain have a smaller value than gaussian noise/defocus blur domains.
Note that, the pre-trained model exhibits a flatter log-likelihood surface for the source data than the target data since the pre-trained parameters are already optimized on the source domain.
From this, we conjecture that our method is capable of identifying the importance of layers by referring to the surface information of the log-likelihood.

\subsection{Layer-wise diagonal of FIM in each domain}\label{sectionB4}
In our proposed method, layer-wise learning weights are calculated as the trace of the FIM. To analyze this, in Figure~\ref{fig:fim_diag}, we visualize the diagonal of the FIM in gaussian noise domain and defocus blur domain.
In comparison to the source domain, we demonstrate that the lower diagonal values of the FIM within layer $\theta^l$ signify the convergence of that layer, offering advantages in representing a particular domain distribution. Since these diagonal elements reflect the curvature of each layer's distribution with respect to the log-likelihood of model outputs, we can establish that our method efficiently leverages layer sharpness in layer-wise learning. Therefore our method, which selects layers to optimize using Hessian approximation with FIM, can effectively employ auto-weighting to identify layers for preservation or concentrated adaptation.
\end{document}